\documentclass{article}

     \PassOptionsToPackage{numbers, compress}{natbib}

 \usepackage[preprint]{neurips_2023}

\usepackage[utf8]{inputenc} 
\usepackage[T1]{fontenc}    
\usepackage{hyperref}       
\usepackage{url}            
\usepackage{booktabs}       
\usepackage{amsfonts}       
\usepackage{nicefrac}       
\usepackage{microtype}      
\usepackage{xcolor}         
\usepackage{CJKutf8}
\usepackage{graphicx}
\usepackage{amsmath}
\usepackage{multirow}

\newcommand{\model}{\textsc{Taiwan-LLM}}

\title{\model: Bridging the Linguistic Divide with \\a Culturally Aligned Language Model}

\author{%
    Yen-Ting Lin \quad Yun-Nung Chen\\
    Department of Computer Science and Information Engineering\\
    National Taiwan University\\
  \texttt{\{ytl, y.v.chen\}@ieee.org} \\
  \url{https://twllm.com}
}

\begin{document}

\maketitle

\begin{abstract}
In the realm of language models, the nuanced linguistic and cultural intricacies of Traditional Chinese, as spoken in Taiwan, have been largely overlooked. This paper introduces \model, a pioneering Large Language Model that specifically caters to the Traditional Chinese language, with a focus on the variant used in Taiwan. Leveraging a comprehensive pretraining corpus and instruction-finetuning datasets, we have developed a model that not only understands the complexities of Traditional Chinese but also embodies the cultural context of Taiwan. \model~represents the first of its kind, a model that is not only linguistically accurate but also culturally resonant with its user base. Our evaluations demonstrate that \model~achieves superior performance in understanding and generating Traditional Chinese text, outperforming existing models that are predominantly trained on Simplified Chinese or English. The open-source release of \model~invites collaboration and further innovation, ensuring that the linguistic diversity of Chinese speakers is embraced and well-served. The model, datasets, and further resources are made publicly available to foster ongoing research and development in this field.\footnote{The model is publicly available at \url{https://github.com/MiuLab/Taiwan-LLM}.}
\end{abstract}


\section{Introduction}

The linguistic diversity of the Chinese language, with its various dialects and written forms, presents unique challenges in natural language processing. Traditional Chinese, used predominantly in Taiwan, Hong Kong, and Macau, has been notably underrepresented in the development of language models. This oversight has resulted in a technological divide, where speakers of Traditional Chinese lack access to high-quality language models that understand their language nuances and cultural context. To address this gap, we introduce \model, the first Large Language Model (LLM) designed specifically for the Traditional Chinese language as used in Taiwan.

Recent advancements in LLMs have seen models like GPT-4 and Claude2 achieve remarkable performance in understanding and generating text. However, these models are predominantly trained on English and Simplified Chinese (zh-cn), leading to a misalignment with the linguistic intricacies of Traditional Chinese (zh-tw). This misalignment not only affects the accuracy of language generation but also overlooks the rich cultural heritage embedded within the language, which is crucial for applications such as digital humanities and culturally sensitive communication.

In this work, we focus on developing an LLM that is both linguistically and culturally aligned with Traditional Chinese speakers in Taiwan. We leverage a large-scale pretraining corpus that encompasses a wide range of text genres, from literature to colloquial dialogues, ensuring that our model captures the breadth of Traditional Chinese language use. Additionally, we employ instruction-finetuning datasets that incorporate cultural nuances and regional idiomatic expressions, further enhancing the model's alignment with Taiwanese users' expectations.

\model~sets a new precedent for language models by providing a culturally resonant tool for the Traditional Chinese-speaking community. Our evaluations demonstrate that \model~not only excels in standard language understanding and generation tasks but also shows a deep comprehension of cultural references and social norms specific to Taiwan. This breakthrough is a step towards bridging the linguistic divide and offering equitable access to language technologies for all Chinese speakers.

The contributions of this paper are manifold: we present the first LLM that is fine-tuned to the linguistic and cultural context of Traditional Chinese as spoken in Taiwan; we provide a comprehensive evaluation that showcases the model's superior performance over existing models; and we release our model, datasets, and resources to the public, fostering further research and development in this area. Models, code, and instructions are available at \url{https://github.com/MiuLab/Taiwan-LLaMa}.

\section{Related Work}

The landscape of open large language models (LLMs) has expanded rapidly, providing valuable resources for the research community. The release of models like ChatGPT and LLaMA~\citep{Touvron2023-af} has catalyzed a wave of research into efficient fine-tuning, context management, and generative capabilities. Subsequent models such as MosaicML's MPT~\citep{mpt}, Together AI's RedPajama-INCITE~\citep{incite}, TII's Falcon~\citep{penedo2023refinedweb}, Meta's Llama 2~\citep{Touvron2023-af}, and Mistral 7B~\citep{Jiang2023-qi} have continued this trend, each contributing unique features and improvements. Our work with \model~builds upon these developments, particularly leveraging the strong performance of Llama 2 as a foundation.

The evolution of LLMs has also seen a focus on enhancing small model performance through techniques like distillation from larger counterparts. Initiatives such as the self-instruct method~\citep{Wang2023-ka}, Alpaca~\citep{Taori2023-ga}, and Vicuna~\citep{Chiang2023-rz} have paved the way for refined distillation processes. While these efforts have concentrated on distilling the supervised fine-tuning (SFT) stage, our approach encompasses both SFT and preference optimization. Notably, models like WizardLM~\citep{Xu_undated-us} and Xwin-LM~\citep{Team2023-fp} have explored advanced methods beyond dSFT, including distilling preference optimization through PPO~\citep{Schulman2017-mu}, which we benchmark against in our evaluations.

As LLMs have grown more powerful, so have the tools for their evaluation. Benchmarks like the LMSYS chatbot arena~\citep{Zheng2023-yf}, AlpacaEval~\citep{dubois2023alpacafarm}, and LLM-Eval~\citep{lin-chen-2023-llm} utilize either human judgment or other LLMs like GPT-4 and Claude to assess model outputs. MTBench~\citep{Zheng2023-yf} represents another step forward, employing GPT-4 to score responses across a variety of tasks. Additional evaluation platforms include the HuggingFace Open LLM leaderboard~\citep{open-llm-leaderboard}, the Chain-of-Thought Hub~\citep{fu2023chainofthought}, and FastEval~\citep{fasteval}. Our work with \model~is evaluated on TC-Eval~\citep{hsu2023advancing} to ensure a comprehensive assessment of its performance.

\section{Method}

\begin{figure}
    \centering
    \includegraphics[width=1.0\linewidth]{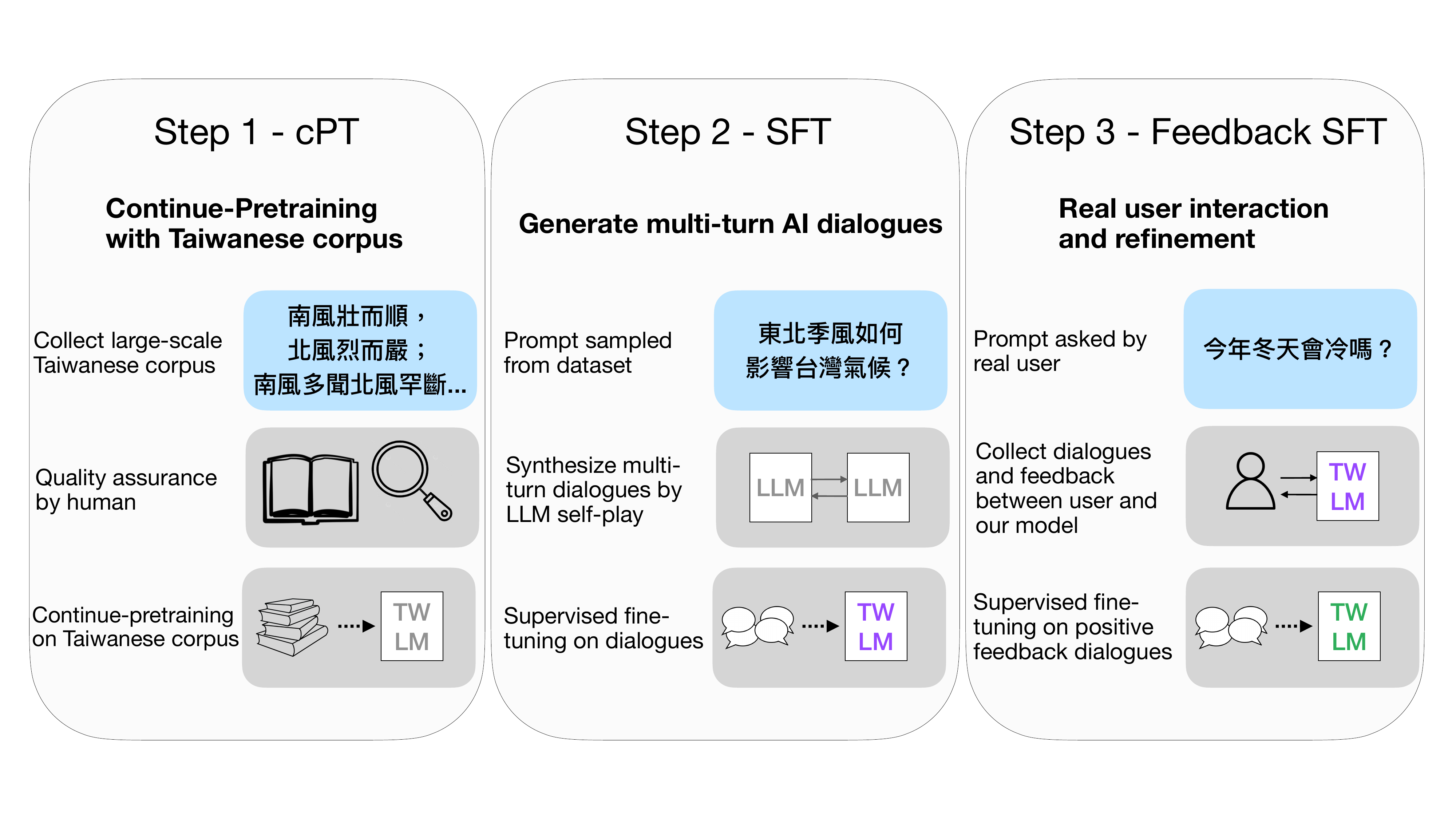}
    \vspace{-10mm}
    \caption{The three-phase methodology for \model~development: (1) cPT - Continue-Pretraining on a large-scale Taiwanese corpus with quality assurance checks, (2) SFT - Supervised Fine-Tuning on multi-turn dialogues through prompt datasets and LLM self-play, and (3) Feedback SFT - Enhancing model performance through real user interactions and subsequent refinement loops, leveraging native speaker insights for cultural and linguistic accuracy.}
    \label{fig:taiwanllm_process}
\end{figure}

The objective of this work is to create a language model that is finely attuned to the linguistic subtleties and cultural context of Traditional Chinese as used in Taiwan. Our approach, depicted in Figure~\ref{fig:taiwanllm_process}, consists of three main stages: Continue-Pretraining (cPT), Supervised Fine-Tuning (SFT), and Feedback Supervised Fine-Tuning (Feedback SFT).

\paragraph{Continue-Pretraining (cPT)}

In the continue-pretraining phase, we start with a base language model, say $\pi_{\text{base}}$, such as Llama2, and further pretrain it on a large-scale Taiwanese corpus, denoted as $\mathcal{C}_{\text{TW}}$. The objective of this phase is to maximize the log-likelihood of the Taiwanese corpus under the model, which can be expressed mathematically as:

\[ \pi_{\text{cPT}} = \arg\max_{\pi} \mathop{\mathbb{E}}_{(x, y) \sim \mathcal{C}_{\text{TW}}} \log  \pi(y \mid x) \]

where $x$ represents a sequence of tokens and $y$ is the next token to predict.

\paragraph{Supervised Fine-Tuning (SFT)}

After continue-pretraining, we fine-tune the model on a dataset of multi-turn dialogues, denoted as $\mathcal{D}_{\text{dialogue}}$. This dataset consists of pairs of prompts and responses, and the goal is to adjust the model parameters to maximize the likelihood of the responses given the prompts. This can be expressed as:

\[ \pi_{\text{SFT}} = \arg\max_{\pi} \mathop{\mathbb{E}}_{(x, y) \sim \mathcal{D}_{\text{dialogue}}} \log  \pi(y | x) \]

where $x$ represents a prompt and $y$ is the corresponding response. 

The SFT phase helps the model to better understand the nuances of conversation and to generate more coherent and engaging responses. It's worth noting that the SFT phase is performed on top of the continue-pretrained model, $\pi_{\text{cPT}}$, thus leveraging the broad language understanding acquired in the cPT phase while also honing the model's ability to generate conversational responses.

\paragraph{Feedback Supervised Fine-Tuning (Feedback SFT)}

In the final stage, we refine the model based on positive feedback from real users. We collect this feedback through a user interface that allows users to interact with the model and provide binary ratings (positive or negative) on its responses. We only keep the instances where the rating is positive for feedback SFT.

This feedback can be represented as a dataset $\mathcal{F}_{\text{pos}}$ of pairs $(x, y)$, where $x$ is a prompt, $y$ is the model's response, and the user's rating of the response is positive. The goal of feedback SFT is to adjust the model parameters to maximize the likelihood of the positively-rated responses given the prompts, which can be expressed as:

\[ \pi_{\text{Feedback SFT}} = \arg\max_{\pi} \mathop{\mathbb{E}}_{(x, y) \sim \mathcal{F}_{\text{pos}}} \log  \pi(y | x) \]

This phase is performed on top of the supervised fine-tuned model, $\pi_{\text{SFT}}$, thus leveraging the conversational abilities acquired in the SFT phase while also incorporating user feedback to better align the model with user preferences.

In future work, we plan to explore more sophisticated training methods such as Reinforcement Learning from Human Feedback (RLHF) and Direct Preference Optimization (DPO) to further improve the model's performance based on user feedback.

\section{Experiments}

Our fine-tuning experiments are conducted using Llama2~\citep{Jiang2023-qi}, a state-of-the-art base language model that has demonstrated strong performance on various NLP benchmarks. We utilize the Transformer Reinforcement Learning (TRL) library for training~\citep{Von_Werra2020-yq}, coupled with DeepSpeed ZeRO-2~\citep{deepspeed} and FlashAttention-2~\citep{flash_attention2} to optimize memory usage and enhance training speed. All models are trained using the AdamW optimizer without weight decay. All experiments were conducted on up to 48 H100s using bfloat16 precision.

\subsection{Datasets}

We utilize three primary datasets for the development of \model:

\begin{enumerate}
    \item \textbf{Continue-Pretraining Corpus:} As detailed in Table~\ref{tab:corpus}, our continue-pretraining corpus is a comprehensive collection of documents from various sources, meticulously curated to represent the Taiwanese context. Each source of data is verified by the authors for quality assurance to primarily avoid spam and toxic content. This verification is conducted at the website level rather than the document level, allowing for a more efficient filtering process while maintaining a high standard of data quality. Future work may involve more granular document-level analysis to further enhance the quality of the training corpus.
    

    \item \textbf{Supervised Fine-Tuning (SFT) Data:} For SFT, we compile a diverse set of instruction datasets, as shown in Table~\ref{tab:ft-data}. We translate prompts from various instruction datasets using \texttt{gpt-3.5-turbo} and then generate responses to these prompts with the same model. Additionally, the authors have created a unique dataset, \textit{Taiwan Instruction}, based on the experiences with the pilot model. This dataset consists of author-written conversations that are specifically designed to capture the cultural and linguistic nuances of Taiwan, ensuring that the model can handle locally relevant scenarios and idiomatic expressions.
    
    \item \textbf{Feedback Supervised Fine-Tuning (Feedback SFT) Data:} We collect 20,000 user feedback instances from a dedicated platform (https://twllm.com) and incorporate this data into the SFT dataset. This feedback is used to further refine the model, ensuring that it aligns with user preferences and expectations.
\end{enumerate}

The continue-pretraining corpus is designed to provide a broad understanding of the language, while the SFT and Feedback SFT datasets are intended to refine the model's ability to engage in dialogue and respond to instructions accurately.

\begin{table}[h!]
  \centering
  \caption{The \model~continue-pretraining Corpus.}
  \label{tab:corpus}
  \begin{tabular}{lrrr}
    \toprule
    \textbf{Data source} & \textbf{Documents} & \textbf{Tokens} & \textbf{Token \%} \\
    \midrule
    Social media & 8.24 millions & 16.6 billions & 47.32\% \\
    News & 8.60 millions & 10.4 billions & 29.56\% \\
    Knowledge base & 3.19 millions & 5.7 billions & 16.29\% \\
    Books & 4 thousands & 2.4 billions & 6.83\% \\
    \midrule
    Total & 20.0 millions & 35.1 billions & 100.00\% \\
    \bottomrule
  \end{tabular}
\end{table}

\begin{table}
\small
\centering
\caption{Supervised fine-tuning datasets used in \model.}
\label{tab:ft-data}
\begin{tabular}{llr}
\toprule
\bf Datasets & \bf Prompts Sourced from & \bf Instances \\
\midrule
~~SuperNI~\cite{wang2022super} & NLP datasets + Human-written Instructions & 18,547\\
~~CoT~\cite{wei2022chain} & NLP datasets + Human-written CoTs & 35,990\\
~~Flan V2~\cite{longpre2023flan} & NLP datasets + Human-written Instructions & 16,782\\
~~Dolly~\cite{conover2023free} & Human-written from scratch & 14,752\\
~~Open Assistant 1~\cite{kopf2023openassistant} & Human-written from scratch & 14,797\\
~~Self-instruct & Generated w/ vanilla GPT3 LM & 48,409\\
~~Unnatural Instructions & Generated w/ Davinci-002 & 30,268\\
~~Alpaca & Generated w/ Davinci-003 & 41,133\\
~~Code-Alpaca~\cite{chaudhary2023code} & Generated w/ Davinci-003 & 20,111\\
~~GPT4-Alpaca~\cite{peng2023instruction} & Generated w/ Davinci-003 + GPT4 & 22,472\\
~~Baize & Generated w/ ChatGPT & 67,699\\
~~ShareGPT\footnote{ShareGPT (\url{https://sharegpt.com/}) data was used to build the Vicuna model~\cite{chiang2023vicuna}. We instead use a reproduced version from \url{https://huggingface.co/datasets/anon8231489123/ShareGPT_Vicuna_unfiltered/tree/main/HTML_cleaned_raw_dataset}, and follow Vicuna to split the long conversations into blocks with a maximum length of 2048 tokens} & User prompts + outputs from various models & 79,762\\
~~Evol Instruction & Generated w/ ChatGPT & 33,798\\
~~Airoboros & Generated w/ ChatGPT & 39,071 \\
\midrule
~~Bilingual News Corpus & Traditional Chinese-English Parallel News Texts & 13,104 \\
~~Taiwan Instruction & Author-written Conversations & 947\\
~~twllm.com Feedback & User-written Conversations & 20,00 \\
\bottomrule
\end{tabular}
\end{table}

\subsection{Evaluation}
For the evaluation of \model, we leverage the TC-Eval benchmark suite~\citep{hsu2023advancing}, which provides a comprehensive set of tasks tailored for assessing the capabilities of Traditional Chinese language models. The suite includes benchmarks for contextual question-answering, summarization, classification, and table understanding. While we adopt the same metrics used in the TC-Eval study, such as Exact Match (EM) for question-answering tasks and ROUGE-2 for summarization, we adapt the metrics from the TC-Eval study to better suit the free-form responses generated by LLMs.

We assess \model's performance on several datasets:

\textbf{Contextual QA:} We use DRCD, a Traditional Chinese machine reading comprehension dataset, and FGC, which contains questions from Taiwanese news articles and government announcements.

\textbf{World Knowledge:} We evaluate the model's common sense abilities and knowledge of Taiwanese culture with the TTQA dataset and test its problem-solving skills with the TMMLU dataset.

\textbf{Summarization:} The model's ability to generate abstractive summaries is measured using the XSum-TC dataset.

\textbf{Classification:} We use the IMDB-TC dataset to evaluate sentiment classification capabilities.

\textbf{Table Understanding:} The model's understanding of tabular data is tested with the Penguins-in-a-Table-TC task.

Metrics for evaluation include Exact Match (EM) for tasks requiring precise answers and ROUGE-2 for summarization quality assessment.

\model~is also compared against a range of models, including those specifically fine-tuned for the Traditional Chinese language as well as larger, more general models. This comparison helps to contextualize \model's performance relative to the current state-of-the-art.

\section{Results and Ablations}

\begin{table}[h!]
  \centering
  \caption{Performance of \model~and baselines on TC-Eval.}
  \label{tab:performance}
  \resizebox{\columnwidth}{!}{%
  \begin{tabular}{lcccccccc}
    \toprule
    \multirow{2}{*}{\textbf{Model}} & \textbf{DRCD} & \textbf{FGC} & \textbf{TTQA} & \textbf{TMMLU} & \textbf{Xsum} & \textbf{IMDB} & \textbf{Table} & \multirow{2}{*}{\textbf{Avg}} \\
    & (EM) & (EM) & (Acc) & (Acc) & (Rouge2) & (Acc) & (Acc) & \\
    \midrule
    \model~13B & 87.57\% & 50.00\% & 70.87\% & 39.04\% & 5.23\% & 92.36\% & 32.89\% & \textbf{53.99\%} \\
    \quad - cPT & 75.81\% & 38.00\% & 56.31\% & 36.28\% & 0.06\% & 93.94\% & 26.84\% & 46.75\% \\
    \quad + CoomonCrawl & 70.08\% & 34.00\% & 77.67\% & 31.53\% & 3.92\% & 79.36\% & 26.17\% & 46.11\% \\
    \midrule
    \model~7B & 84.11\% & 46.00\% & 54.37\% & 30.03\% & 4.60\% & 86.04\% & 28.19\% & \textbf{47.62\%} \\
    \quad - Feedback SFT & 84.51\% & 42.00\% & 55.34\% & 32.01\% & 5.21\% & 87.50\% & 25.50\% & 47.44\% \\
    \quad + CommonCrawl & 69.80\% & 36.00\% & 48.54\% & 27.49\% & 4.37\% & 59.82\% & 20.13\% & 38.02\% \\
    \midrule
    \texttt{GPT-4} & 96.68\% & 42.00\% & 53.40\% & 60.48\% & 4.30\% & 86.90\% & 62.42\% & 58.03\% \\
    \texttt{GPT-3.5 turbo} & 90.75\% & 42.00\% & 55.34\% & 55.61\% & 3.71\% & 91.38\% & 38.93\% & 53.96\% \\
    \texttt{Claude-2.1} & 93.56\% & 40.00\% & 68.93\% & 59.86\% & 3.57\% & 94.52\% & 48.99\% & \textbf{58.49\%} \\
    \texttt{Claude-instant-1.2} & 78.99\% & 34.00\% & 68.93\% & 55.28\% & 4.21\% & 91.14\% & 47.65\% & 54.31\% \\
    \midrule
    \texttt{Llama-2-13b-chat} & 45.95\% & 18.00\% & 59.22\% & 35.36\% & 0.00\% & 52.40\% & 27.52\% & 34.06\% \\
    \texttt{Llama-2-7b-chat} & 28.60\% & 12.00\% & 50.49\% & 29.34\% & 0.00\% & 51.04\% & 18.12\% & 27.08\% \\
    \bottomrule
  \end{tabular}
  }
\end{table}

The results of our experiments, as shown in Table~\ref{tab:performance}, indicate that \model, with its 13 billion parameters, achieves competitive performance when compared to proprietary models such as GPT-4 and Claude-2.1. Notably, \model's average performance across all tasks is 53.99\%, which is comparable to GPT-3.5 turbo's 53.96\%. This demonstrates the effectiveness of our approach in creating a model that is well-suited for Traditional Chinese, despite having fewer parameters than some proprietary models.

\subsection{Impact of Continue-Pretraining (cPT)}
Our ablation study reveals the significant impact of continue-pretraining (cPT) on \model's performance. When we remove the cPT phase, there is a noticeable drop in performance across most tasks. For instance, the EM score on the DRCD dataset decreases from 87.57\% to 75.81\%, and the accuracy on the TTQA dataset drops from 70.87\% to 56.31\%. This highlights the importance of cPT in adapting the model to the linguistic characteristics of Traditional Chinese.

\subsection{Impact of Feedback Supervised Fine-Tuning (Feedback SFT)}
We also investigate the effect of incorporating user feedback through Feedback SFT. The results show that this phase contributes to the model's performance, albeit to a lesser extent than cPT. For example, the 7 billion parameter version of \model~without Feedback SFT achieves an average performance of 47.44\%, only slightly lower than the 47.62\% with Feedback SFT. This suggests that while Feedback SFT helps in fine-tuning the model to user preferences, the core capabilities are largely established during the cPT and SFT phases.

\subsection{Impact of Adding Web Data}
When we added approximately 9 billion tokens of CommonCrawl data, specifically filtered for Traditional Chinese content (zh-tw), the performance of \model~unexpectedly declined. For the 13 billion parameter model, the Exact Match (EM) score on the DRCD dataset decreased from 87.57\% to 70.08\%, and the accuracy on the TTQA dataset saw a slight increase to 77.67\%, indicating a complex impact. The 7 billion parameter model experienced a more significant drop, with the average performance across all tasks falling from 47.62\% to 38.02\%.

This reduction in performance, despite the substantial volume of additional tokens, underscores the critical role of data quality over sheer quantity. The introduction of web crawl data, which may contain a mix of high-quality and lower-quality content, can introduce noise that detracts from the model's ability to accurately process and generate Traditional Chinese text. This result highlights the necessity of meticulous data curation to ensure the training material aligns with the linguistic and cultural nuances of the target language.

\subsection{Comparison with Open-Source Models}
When compared to the original Llama-2 models, which have not been specifically trained on Traditional Chinese, \model~demonstrates a clear advantage. The original Llama-2-13b-chat model achieves an average performance of 34.06\%, significantly lower than \model's 53.99\%. This underscores the necessity of targeted training on Traditional Chinese to ensure high-quality performance for users in Taiwan.

\subsection{Comparison with Proprietary Models}
Our 13 billion parameter \model~is on par with proprietary models like GPT-3.5 turbo, which is a testament to the effectiveness of our culturally and linguistically tailored training approach. While proprietary models like GPT-4 and Claude-2.1 still lead in some areas, \model's competitive performance suggests that with continued refinement and expansion of training data, open-source models can achieve parity with or even surpass proprietary models for specific languages and regions.

The results confirm that \model~effectively supports Traditional Chinese, thanks to our specialized continue-pretraining and fine-tuning for Taiwanese usage. Its release is a major advancement for underrepresented language communities.

\subsection{Qualitative Study}
To enhance the investigation of whether \model~exhibits a superior grasp of Taiwanese culture, we manually created a compact collection of questions deeply rooted in Taiwanese cultural context. 
These questions were designed to probe the model's ability to interpret and respond in a manner that reflects an intrinsic understanding of local nuances and cultural specificities.
The selected examples, as presented in Table~\ref{tab:qualitative}, offer a clear illustration of this capability. 
In these instances, \model~demonstrates a remarkable alignment with the intricacies of Taiwanese culture, distinguishing itself through responses that not only answer the queries accurately but also resonate with the cultural context of Taiwan.
This qualitative analysis showcases the model's adeptness in navigating cultural nuances, reinforcing its effectiveness in bridging the linguistic and cultural divide in a Taiwanese context.

\begin{CJK*}{UTF8}{bkai}
\begin{table}[t!]
\small
  \centering
  \caption{The samples requiring culture understanding.}
  \label{tab:qualitative}
  \begin{tabular}{p{2.8cm}p{10cm}}
    \toprule
    \textbf{Question} & \textbf{Answer}\\
    \midrule
    & \bf ChatGPT\\
    & NTU 位於新加坡，具體地址是50 Nanyang Avenue, Singapore 639798。 \\
    NTU 在哪? & \bf \model\\
     & NTU 是國立臺灣大學的縮寫。它是臺灣的一所頂尖研究型大學，位於\underline{臺北市}。\\
    \midrule
    & \bf ChatGPT\\
    & "22K" 可能指的是金的純度，尤其是在珠寶和金飾方面。\\
    甚麼是22K?& \bf \model \\
    & "22K"是指\underline{台灣的大學畢業生的起薪}，大約是新台幣22,000元。這個詞被用來形容大學畢業生的起薪偏低，並且在就業市場上對於大學畢業生的需求不足。\\
    \midrule
    & \bf ChatGPT\\
    買完東西店員詢問：「您要使用載具嗎？」這是甚麼意思呢？& 當店員詢問「您要使用載具嗎？」時，他們實際上是在問您是否需要提供一個購物袋或容器，以便將您購買的物品放入其中，以便您更容易攜帶回家或其他地方。這種問法通常出現在一些國家或地區，特別是考慮到環保因素的情況下，為了減少使用一次性塑料袋，商家會提供可重複使用的購物袋或容器，或者鼓勵顧客自己攜帶可重複使用的袋子或容器。如果您需要購物袋或容器，您可以回答「是的」，店員通常會為您提供相應的選項。如果您不需要，則可以回答「不需要，謝謝」。\\
    & \bf \model \\
    & 在台灣，我們常見到一些店家會提供消費者使用載具的選項。載具指的是可以存取電子發票的設備，例如手機或平板電腦。當店員問您「您要使用載具嗎？」時，他們想知道您是否希望\underline{將發票存放在自己的載具上，而不是列印出來}。如果您選擇使用載具，店員會將發票直接發送到您的載具中，您可以隨時查看或列印出來。使用載具有幾個好處，包括減少紙張浪費和方便查看發票。此外，使用載具還可以參加政府主辦的抽獎活動，有機會贏得大獎！\\
    \bottomrule
  \end{tabular}
\end{table}
\end{CJK*}

\section{Conclusion}
In conclusion, \model~represents a significant stride forward in the development of language models tailored to the linguistic and cultural nuances of Traditional Chinese as spoken in Taiwan. Our comprehensive approach, encompassing continue-pretraining on a curated Taiwanese corpus, supervised fine-tuning with instruction datasets, and refinement through user feedback, has culminated in a model that not only understands the complexities of the language but also resonates with the cultural context of its speakers. The evaluations demonstrate that \model~achieves competitive performance, particularly when compared to larger, proprietary models trained on more generalized datasets. This underscores the importance of specialized training for language models to effectively serve specific linguistic communities.

The open-source release of \model~invites collaboration and innovation, setting a precedent for future research in language technology. By providing a model that is both linguistically accurate and culturally aligned, we aim to bridge the gap in language technology for Traditional Chinese speakers and contribute to the global effort of ensuring equitable access to high-quality language models. As we continue to refine \model~and expand its capabilities, we anticipate that it will not only serve as a valuable tool for the Taiwanese community but also inspire similar initiatives for other underrepresented languages around the world.

\section*{Acknowledgements}
We thank Ubitus K.K. for the GPU compute grant.

\section*{Limitation}
We acknowledge that while our model represents a significant advancement in language technology for Traditional Chinese speakers, it is the first step in an ongoing journey. Future work will need to address the continuous evolution of language and culture, ensuring that \model~remains a relevant and valuable resource for its users.

\bibliographystyle{acl_natbib}
\bibliography{custom}

\end{document}